\documentclass[conference]{IEEEtran}
\ifCLASSINFOpdf
  \usepackage[pdftex]{graphicx}
  
\else
\fi
\usepackage{amsthm}
\usepackage{amsmath}
\usepackage{amssymb}
\usepackage{array}
\usepackage[T1]{fontenc}
\usepackage{tabularx}
\usepackage{enumerate}
\usepackage{hyperref}
\usepackage{cite}
\begin{document}

%\title{Scalable Representation of Massive Hierarchical Data for Probabilistic-based Problem Solving}
\title{Entity Type Recognition using \\an Ensemble of
Distributional Semantic Models \\to Enhance Query
Understanding}



	
	\author{\IEEEauthorblockN{Walid Shalaby\IEEEauthorrefmark{1},
			Khalifeh Al Jadda\IEEEauthorrefmark{3},
			Mohammed Korayem\IEEEauthorrefmark{3} and
			Trey Grainger\IEEEauthorrefmark{3}}
		\IEEEauthorblockA{\IEEEauthorrefmark{1}Computer Science Department\\
			University of North Carolina at Charlotte,\\
			Charlotte, NC, USA\\ Email: wshalaby@uncc.edu}
		\IEEEauthorblockA{\IEEEauthorrefmark{3}CareerBuilder\\
			Norcross, GA, USA\\
			Email: mohammed.korayem, khalifeh.aljadda, trey.grainger@careerbuilder.com}}
		

\maketitle

\begin{abstract}
%\boldmath
We present an ensemble approach for categorizing search query entities in the recruitment domain. Understanding the types of entities expressed in a search query ({\it Company, Skill, Job Title, etc.}) enables more intelligent information retrieval based upon those entities compared to a traditional keyword-based search. Because search queries are typically very short, leveraging a traditional bag-of-words model to identify entity types would be inappropriate due to the lack of contextual information. Our approach instead combines clues from different sources of varying complexity in order to collect real-world knowledge about query entities. We employ distributional semantic representations of query entities through two models: 1) contextual vectors generated from encyclopedic corpora like Wikipedia, and 2) high dimensional word embedding vectors generated from millions of job postings using word2vec. Additionally, our approach utilizes both entity linguistic properties obtained from WordNet and ontological properties extracted from DBpedia. We evaluate our approach on a data set created at CareerBuilder; the largest job board in the US. The data set contains entities extracted from millions of job seekers/recruiters search queries, job postings, and resume documents. After constructing the distributional vectors of search entities, we use supervised machine learning to infer search entity types. Empirical results show that our approach outperforms the state-of-the-art word2vec distributional semantics model trained on Wikipedia. Moreover, we achieve micro-averaged $F1$ score of 97\% using the proposed distributional representations ensemble.
\end{abstract}


% \fi
\IEEEpeerreviewmaketitle

\section{Introduction}
Entity Recognition ({\it ER}) is an information extraction task which refers to identifying regions of text corresponding to entities. A sub-task related to {\it ER} is the Entity Type Recognition ({\it ETR}) which refers to categorizing these entities into a predefined set of types~\cite{kazama2007exploiting}. The focus of the majority of {\it ETR} research has been on Named Entity Recognition ({\it NER}), which typically limits entity types to $Person$, $Location$, and $Organization$~\cite{nadeau2007survey,minkov2005extracting,shaalan2007person,zhou2002named}.  Most techniques used in {\it ETR} rely on a mix of local information about the context of the entity and external knowledge usually gained through learning on training data. {\it ETR} in search queries is considered extremely important; a Microsoft's study reported that $71\%$ of queries submitted to their Bing search engine contain named entities somewhere, while $20-30\%$ are purely named entities~\cite{yin2010building}. Recognizing the type of named entities in queries enables a search engine to understand the intent of users, which subsequently leads to more accurate results being returned. {\it ETR} in search queries is very challenging, however, due to the lack of textual context surrounding the query. Search queries are usually made of just a few words, which is typically not enough context to independently and accurately recognize the types of the entities within a search query. 
Our research is specifically targeted at the problem of {\it ETR} within the job search and recruitment domain.  Unfortunately, none of the published {\it ETR} datasets fully resemble the entity categories within the job search and recruitment domain. Some of the specific entity categories within this domain include {\it Company}, {\it Job} {\it Title}, {\it School}, and {\it Skill}, which all aren't found explicitly within existing {\it ETR} datasets. As a result, we can't leverage any existing gazetteers for these entity types.

In this paper we introduce a novel system for {\it ETR} in search queries which has been applied successfully within the job search and recruitment domain. The proposed system utilizes features collected from Wikipedia, DBpedia
%\footnote{\url{http://wiki.dbpedia.org/}}
, WordNet 
%~\cite{wordnet}
, and a corpus of more than 60 million job postings provided by Careerbuilder. We integrated this model within CareerBuilder's semantic search engine~\cite{aljadda2014pgmhd,aljadda2015improving,korayem2015query}, which improved the quality of search results for tens of millions of job seekers every month. 

The system is used within the search engine in two ways: 1) offline, to classify a list of pre-recognized entities extracted from popular queries found in CareerBuilder's search logs, and 2) online, to dynamically classify the search entities within new, previously unseen queries as part of CareerBuilder's semantic query parser.

To the best of our knowledge we are the first group targeting {\it ETR} of queries within the job search and recruitment domain. We evaluated this system using a data set provided by CareerBuilder which contains more than 177K labeled entities. The results demonstrate that our system achieves a $97\%$ micro-averaged $F1$ score over all the categories.

The main contributions of this paper are:
\begin{enumerate}
	\item We introduce a novel approach for generating distributional semantic vectors of named entities in search queries using Wikipedia as an intermediate corpus.
	\item Our approach is simple and efficient. It outperforms state-of-the-art techniques for distributional representations like word2vec.
	\item We evaluate our method on the largest labeled entity type data set within the recruitment domain achieving a 97\% micro-averaged $F1$ score.
	\item We demonstrate increases in overall system accuracy through an ensemble of features leveraging distributional semantic representations, entity ontologies, and entity linguistic properties.
\end{enumerate}
\section{Related Work}

Both {\it ETR} and {\it NER} have experienced a surge in the research community in recent years~\cite{tjong2003introduction,ratinov2009design,carmel2014erd,moro2015semeval,ren2015clustype,milne2008learning,cucerzan2007large}. David et al.~\cite{nadeau2007survey} and Mansouri et al.~\cite{mansouri2008named} presented comprehensive reviews about different approaches for {\it NER} including several representations that leverage dictionaries, corpora, and various classification methods.

Guo et al.~\cite{guo2009named} presented a formulation for both {\it NER} and {\it ETR} in search queries using a probabilistic approach and Latent Dirichlet Allocation ({\it LDA}). They represented query terms as words in documents and modeled the entity type classes as topics. They proposed using a weakly supervised learning algorithm to learn the topics, while impressive, their approach was limited to recognizing only one entity per query. Our approach, instead, can accurately identify multiple entities per search query and recognize their types. 

Other approaches which utilize knowledge bases to link named entities in text with corresponding entities in the knowledge bases were presented in~\cite{kazama2007exploiting,kulkarni2009collective,han2011collective,han2012entity,lin2012entity}.
Wikipedia has been used extensively as a knowledge base for {\it ETR}. Many researchers have utilized Wikipedia-based features such as wikilinks, article titles and categories, and graph representations of the inner links between Wikipedia pages.

Han et al.\cite{kazama2007exploiting} proposed a methodology which relies on having a Wikipedia page whose title is similar to the given entity. After looking up that page, if any, they extracted the category of that entity from the first line in that page. In our case, we couldn't find a Wikipedia page for most of the popular queries we have, for example, {\it java developer} has no corresponding page in Wikipedia. Our methodology can handle such cases by looking in Wikipedia content not titles for the occurrences of that entity and using the context as a representation in order to recognize the entity type.

Richman and Schone proposed a novel system for multilingual {\it NER}~\cite{richman2008mining} . They utilize wikilinks to identify words and phrases that might be entities within text. Once they recognize the entities, they use category links or interlinks to map those entities with English phrases or categories. 

Using Wikipedia concepts as a representation space for query's intent was introduced in \cite{hu2009understanding}. In this paper each intent domain is represented as a set of Wikipedia articles and categories, then each query intent is predicted by mapping the query into the Wikipedia representation space.

The system introduced in \cite{nothman2008transforming} transforms links to Wikipedia articles into named entity annotations by classifying the target articles into the classic named entity types {\it Person}, {\it Location}, and {\it Organization}.

Utilizing Wikipedia infobox for {\it ETR} was presented in \cite{mohamedidentifying}. The proposed model classifies entities by matching entity attributes extracted from the relevant article infobox with core entity attributes built from Wikipedia infobox templates.

The system introduced in \cite{gattani2013entity} converted Wikipedia into a structured knowledge base ({\it KB}). In this work, the authors converted Wikipedia graph structure into a taxonomy. This was done by finding a single main lineage, called the primary lineage, for each concept. This {\it KB} is used later to extract, link, and classify entities mentioned in a Twitter stream.

We consider \cite{laclavik2015search} as the most related work to ours. In this work, the authors proposed a system that utilizes Wikipedia as an intermediate corpus to categorize search queries. The system works through two phases; in the first phase, a query is mapped to its relevant Wikipedia pages by searching an index of Wikipedia articles. In the second phase, concepts representing retrieved Wikipedia pages are mapped into categories. Though we also utilize a Wikipedia search index to retrieve articles related to query entities, our approach utilizes totally different features and entity representation to infer entity type.%However, in the recruitment domain a query may have different named entities of different types, so categorizing the entire query to one class will impact the search relevancy negatively. For example, a query "Software Engineer Facebook" has two named entities "Software Engineer" and "Facebook". The entity type of "Software Engineer" is "job title" while the entity type of "Facebook" is "Company".

%Classifying the named entities in a query individually is extremely important to help the search engine understand the intent of the user. In our example the user is looking for the "Software Engineer" job title in "Facebook" company. Our system tackles this problem by recognizing each possible named entity in a query, then classifying each one separately. 
\section{Methodology}
In this section we detail our methodology for recognizing search query entity types. Our approach employs two distributional semantic representations of search entities. Moreover, we utilize ontological properties as well as linguistic properties of search entities to improve overall system performance. The ultimate goal of our system is to categorize a given search entity into one of four categories: \textit{Company}, \textit{Job Title}, \textit{Skill}, and \textit{School}. We do plan to expand these categories in the future, but these four represent the most important to initially target.

\subsection{System Overview}

Prior to performing {\it ETR}, it is of course necessary that we first perform {\it ER} on incoming search queries so that we know the entities for which we are trying to identify an entity type. Our methodology for recognizing known entities and performing Entity Extraction from queries was previously described in \cite{industrialPaper}. In essence, we perform data mining on historical search query logs, perform collaborative filtering to determine which queries are used commonly together across many users, and build a semantic knowledge base containing the entities and related entities found from within the mined search logs.

Based upon this semantic knowledge base, we are able to perform entity extraction on future queries for known entities, but we are missing two important components:
\begin{enumerate}
	\item Identification of entities not found in our semantic knowledge base.
	\item Knowledge of the entity type of each identified entity.
\end{enumerate}

\begin{figure*}[bt]
	\centering
	\includegraphics[width=10cm,height=10cm,keepaspectratio]{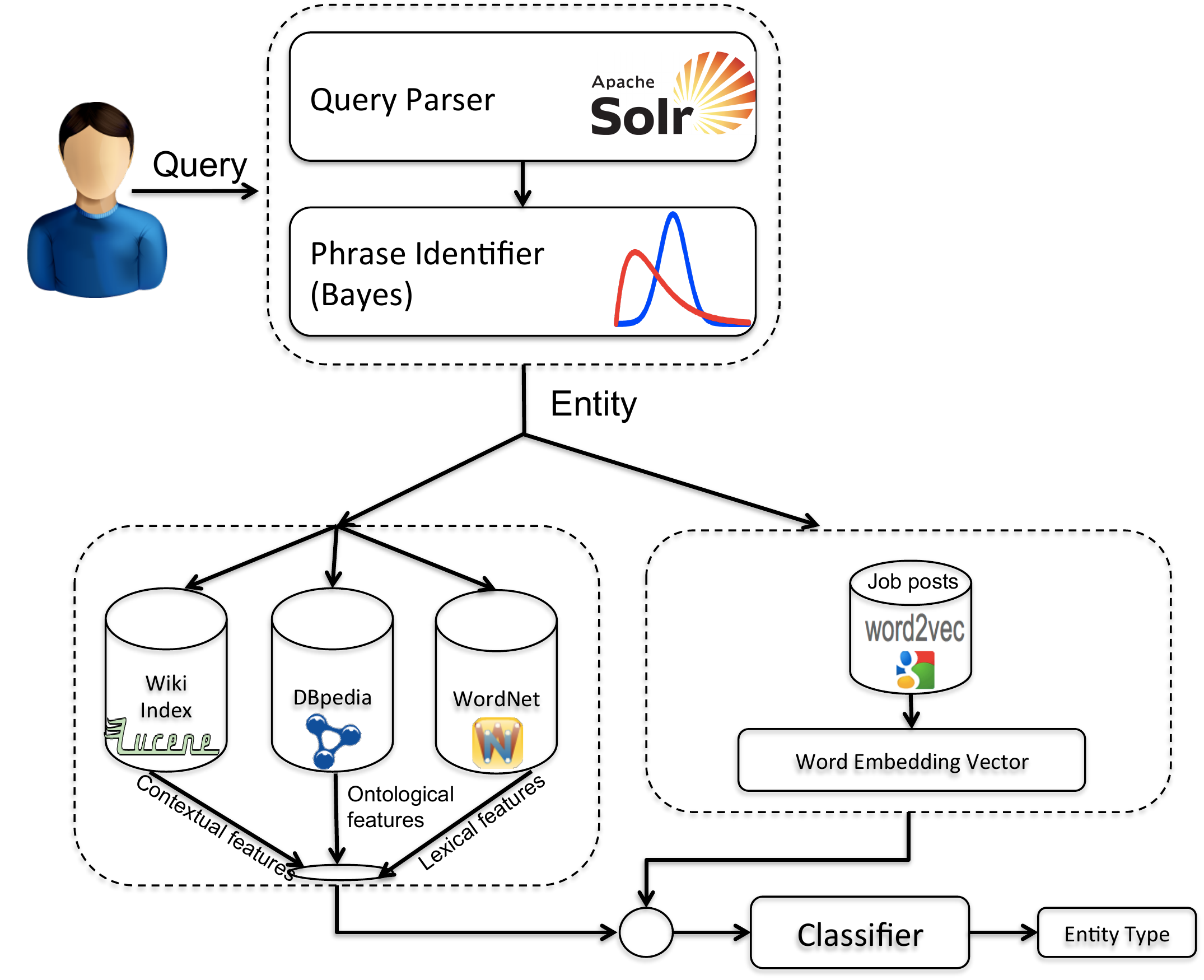}
	\caption{User's query is passed to the query parser and phrase identifier, which perform Entity Resolution leveraging a semantic-knowledge-based and a language-model-based probabilistic parsing. The entities are then enriched using an ensemble of representation models based upon external knowledge base (Wikipedia), %distributional semantic vectors based upon %
		a domain-specific corpus (job postings), ontological features (DBpedia), and Linguistics features (WordNet).}
	\label{system_design}
\end{figure*} 

To solve the first problem, we implemented a language model of unigrams, bigrams, and trigrams across a corpus of millions of job posting documents. Leveraging Bayes algorithm, we are able to dynamically calculate probabilities as to whether any combination of keywords entered into a search query constitute a single phrase or multiple phrases. Based upon the combination of our semantic knowledge base, our Bayes-based phrase identifier, and our query parser, we are able to successfully identify the correct query parsing including the constituent named entities with accuracy of greater than 92\%. 

The last stage needed to truly interpret the user's query correctly is {\it ETR}. If a user searches for {\it google software engineer java}, it is critical to understand that the user is looking for a job at {\it Google (Company)} as a {\it software engineer (Job Title)} programming in {\it Java (Skill)}. Without this knowledge of entity types, we will not be able to fully represent the information need of our users within the search system. The following sections will describe our methodology for performing {\it ETR} on our identified entities.

\subsection{Entity Type Recognition Process}
The proposed system combines features from different sources in order to make accurate entity type predictions for a given search entity. This ensemble of features represents our domain-specific knowledge as well as real-world knowledge about the search entity. We call these features clues. Figure \ref{system_design} shows the system design for how a user's query is parsed, as well as how the system leverages these feature clues to accurately perform {\it ETR}.

The first clue models real-world contextual information about the query entity by searching for that entity inside Wikipedia using a customized search index. The second clue models domain-specific knowledge by building synonym vectors of search entities using the word2vec model \cite{mikolov2013efficient}. These vectors are generated using millions of job postings from CareerBuilder. 

Two other clues, leveraging DBpedia and WordNet, are collected to increase the accuracy and coverage over the \textit{Company} and \textit{Job Title} categories specifically. After collecting all the clues for every known query entity, we combine these features and use them to train an entity classifier over labeled entity samples. The classifier can then be used to categorize new search entities, thus improving our understanding of the query intent for future searches.

\subsection{Constructing Contextual Vectors}
The purpose of this phase is to enrich the contextless search entities with contextual information. In order to do so we map each entity into a distributional semantic vector representation. The vector dimensions represent entity contexts in an intermediate corpus. We use Wikipedia as the source for these contextual vectors for all of the search entities which are represented. 

As query entities need to be categorized in an online fashion, context vectors are required to be constructed as efficiently as possible. Therefore, we build an inverted index of all Wikipedia articles as a preprocessing step. We build the index using Apache Lucene\footnote{https://lucene.apache.org/}, an open-source indexing and search engine. For each article we index the title, content, length, and categories. We exclude all \textit{disambiguation}, \textit{list of}, and \textit{redirect} pages.

As shown in Equation 1, given an entity $e_j$ we construct its context vector $X_{e_j}$ by first searching for that entity in the search index. Then, from the top $n$ search hits, we retrieve all content words $W_i$ that occur in the same context of $e_j$ within a specific window size in each search hit $i$. We also retrieve category words $C_i$ of search hits and add them to $X_{e_j}$.
%\begin{equation}X_{e_j} = {<\!\bigcup_{w \in W_i,w \in C_i} w\!>}:i=[1..n], w \in W_i, c \in C_i\end{equation}
\begin{equation}X_{e_j} \!=\! {<\!w_1,w_2,...,c_1,c_2,...\!>}:w\! \in\! W_i, c\! \in\! C_i, i\!=\![1..n]\end{equation}
%\begin{equation}X_{e_j} = {<w_k,...,w_n, c_k,..c_n>}:  k=1..n, w \in W_i, c \in C_i\end{equation}

These context vectors represent available real-world knowledge about the given entity. Table \uppercase\expandafter{\romannumeral1\relax} shows example search entities along with their context vectors. We can notice that contextual words are very representative for the given entity. Moreover, words from search hits categories augment context words and thus enrich the contextual representation of each entity.

\begin{table*}
	\centering
	\caption{Sample Contextual Vectors}
	%\begin{minipage}{15cm}
		\begin{tabular}{|l|l|}
			\hline
			\multicolumn{2}{|c|}{\textbf{Company}}  \\ \hline
			CareerBuilder     & \textless...market, operate, website, acquired, employment, companies, establishments, ceo...\textgreater \\ \hline
			\multicolumn{2}{|c|}{\textbf{Job Title}}  \\ \hline
			Nurse Assistant  & \textless...journalist, worker, secretary, members, politicians, living, people, youth, office...\textgreater                    \\ \hline
			\multicolumn{2}{|c|}{\textbf{Skill}}  \\ \hline
			Adobe Photoshop    & \textless...editor, graphics, developed, image, file, software, application, version, program...\textgreater          \\ \hline
			\multicolumn{2}{|c|}{\textbf{School}}  \\ \hline
			UNC Charlotte    & \textless...university, north, carolina, college, student, organization, professor, school...\textgreater                 \\ \hline
		\end{tabular}
		%	\footnotetext{Bold words appear within the context window. Italics words are extracted from concept categories. Bold, italics words appear within the context window and are extracted from the concept categories}
	%\end{minipage}
\end{table*}

\subsection{Constructing Synonymy Vectors}
The purpose of this phase is to enrich the search entities with domain-specific knowledge. CareerBuilder has millions of job openings that are posted or modified on daily basis. These postings contain many representative features relevant to the recruitment domain. For example, a typical job posting might contain a job title, job description, required skills, salary information, company information, required experience and education, location...etc. 

In order to utilize this information, we use the job postings as an intermediate corpus to train a word2vec model. For a given search entity $e_j$, we generate its synonyms vector $S_{e_j}$ from words that have closest distributional representations in the trained word2vec model.% As shown in Equation 2, HEREHERE

%\begin{equation}W_{e_j} = \end{equation}

Distributional semantic vectors generated in this phase represent domain-specific knowledge about a given entity. Table \uppercase\expandafter{\romannumeral2\relax} shows the same search entities as in Table \uppercase\expandafter{\romannumeral1\relax} along with corresponding synonymy vectors. We can notice that the \textit{Company} and \textit{School} entity vectors are somewhat poor and unrepresentative. This is because many job postings are missing company information or sometimes company name is only provided without any context. The same problem arises for school information. On the other hand, synonymy vectors of {\it Job Title} and {\it Skill} entities are very rich and representative. This observation motivated us to combine features for search entities from both contextual and synonymy vectors in a combined vector space.

\begin{table*}
	\centering
	\caption{Sample Synonymy Vectors}
	\begin{tabular}{|l|l|}
		\hline
		\multicolumn{2}{|c|}{\textbf{Company}}  \\ \hline
		CareerBuilder     & \textless...us, software, recruiter, digital...\textgreater \\ \hline
		\multicolumn{2}{|c|}{\textbf{Job Title}}  \\ \hline
		Nurse Assistant & \textless...licensed, registered, nurse, rn, lpn, office, coordinator, lvn, midwife...\textgreater                    \\ \hline
		\multicolumn{2}{|c|}{\textbf{Skill}}  \\ \hline
		Adobe Photoshop   & \textless...dreamweaver, flash, acrobat, macromedia, illustrator, pagemaker...\textgreater          \\ \hline
		\multicolumn{2}{|c|}{\textbf{School}}  \\ \hline
		UNC Charlotte       & \textless...raleigh, durham, morrisville, hospital, concord, morrisville, durham...\textgreater                 \\ \hline
	\end{tabular}
\end{table*}

\subsection{Entity Ontological Features}
Another representative feature is extracted from DBpedia by linking search hits (representing Wikipedia concepts) to their corresponding entries in the DBpedia ontology. We use the {\it type} property to determine whether the retrieved concept type is one of our targeted categories, specifically \textit{Company}.

After searching for a given entity $e_j$ in the Wikipedia index, we retrieve the top $n$ search hits (concepts). Then, we check whether the title of any of these concepts is the same as $e_j$. If any, we check whether the type of this concept in DBpedia ontology is \textit{Company} and subsequently add a new binary feature indicating that finding.

Given that companies are already found explicitly in DBPedia, why don't we just use the DBpedia {\it type} feature exclusively for categorizing into the \textit{Company} entity type? There are five reasons we instead choose to combine multiple feature types:
\begin{enumerate}%[i.]
	\item DBpedia ontology suffers from low coverage where many companies in Wikipedia don't have a type of \textit{Company} in DBpedia (e.g., Boonton Iron Works\footnote{\url{https://en.wikipedia.org/wiki/Boonton_Iron_Works}}, SalesforceIQ\footnote{\url{https://en.wikipedia.org/wiki/SalesforceIQ}}).
	\item DBpedia provides categories for the canonical form of company name only. If an entity is searched for using a surface form, the DBpedia lookup will fail. In contrast, Wikipedia will generally contain surface forms in the same context as the canonical form (e.g., International Turnkey Systems Group vs. ITS Group\footnote{\url{https://en.wikipedia.org/wiki/International_Turnkey_Systems_Group}})
	\item As DBpedia covers only Wikipedia concepts, it fails to catch companies that do not have a Wikipedia page. Alternatively, these companies will be correctly categorized using their contextual vectors if mentioned in a representative context within Wikipedia (e.g., Nutonian).
	\item Some companies have a type of \textit{Organization} instead of \textit{Company} in DBpedia. Unfortunately, entities belonging to one of our other entity types ({\it School}) can also be categorized as \textit{Organization} in DBpedia  (e.g., Athens College). This means that we cannot reliably categorize concepts with the type of \textit{Organization} as \textit{Company}.
	\item Finally, there is a time lag between DBpedia and Wikipedia. So, DBpedia does not contain the most recent snapshot of Wikipedia concepts in its ontology.
\end{enumerate}

\subsection{Entity Linguistic Features}
We utilize the lexical properties of search entities to determine whether they belong to one of the target categories, specifically \textit{Job Title}. The motivation behind this approach is the fact that almost all \textit{Job Title} entities contain an agent noun (e.g., director, developer, nurse, manager...etc). To determine whether an entity might represent a \textit{Job Title}, we search its words inside the WordNet dictionary where all agent nouns are stored at the <noun.person> lexical file. Upon finding any, we add a new binary feature indicating that finding.

While it might be tempting to rely exclusively on the agent noun feature from the WordNet lexicon for categorizing \textit{Job Title} entities, two challenges prevent this:
\begin{enumerate}%[i.]
	\item CareerBuilder operates job boards in many countries and in many different languages. Therefore, we're biased toward using language independent models where possible. Depending solely on the WordNet lexicon for categorizing \textit{Job Title} entities would pose limitations on the \textit{ETR} system for non-English job boards.
	\item Not all \textit{Job Title} entities have an agent noun (e.g., staff, faculty).
\end{enumerate}

\subsection{Building the Prediction Model}
To build the \textit{ETR} model, we use supervised machine learning on a very large labeled set of search entities obtained from CareerBuilder's search logs. For each discovered search entity $e_j$, we generate:
\begin{enumerate}
	\item A Contextual vector ($X_{e_j}$) using the Wikipedia index.
	\item A Synonyms vector ($S_{e_j}$) using the word2vec model.
	\item An Ontological type ($ont_{e_j}$) if the entity refers to a DBpedia concept. This is a binary feature which is true if DBpedia type is company.
	\item A Lexical type ($lex_{e_j}$). This is a binary feature which is true if one of the entity terms has a \textit{<noun.person>} type in WordNet, i.e., it is an agent noun.
\end{enumerate}

To combine all those features, we follow a simple yet effective approach. First we utilize the vector space model to generate an entity-word matrix using the distributional semantic vectors ($X_{e_j},S_{e_j}$). The generated distributional vectors represent semantically-related words to the identified query entities, so it is straightforward to then map each entity as a document of words contained in the entity's contextual and synonymy vectors. Rows in the entity-word matrix represent entities and columns represent corresponding related words. Secondly, we transform this matrix using term frequency-inverse document frequency (tf-idf) weights. Thirdly, we append $ont_{e_j}$ and $lex_{e_j}$ as two additional binary columns to the tf-idf entity-word matrix. Finally, we train an entity type classifier on the produced matrix to generate the \textit{ETR} model.

\begin{table}[]
	\centering
	\caption{Distribution of Entities over Categories}
	\label{my-label}
	\begin{tabular}{|l|l|}
		\hline
		\textbf{Category}  & \textbf{Number of instances} \\ \hline
		Company   & 42,934               \\ \hline %42934 
		Job Title & 3,608                \\ \hline %3608 
		School    & 106,153             \\ \hline %106153  
		Skill     & 25,093                \\ \hline %25093
	\end{tabular}
\end{table}

\begin{table*}[]
	\centering
	\caption{Performance of contextual vectors \textit{ETR} model ($wiki_x$) on labeled entities compared to baseline models using 10-fold cross-validation. ($bow$) is the bag-of-words model, ($wiki_w$) is word2vec trained on Wikipedia.}
	\label{my-label}
	\begin{tabular}{|l|c c c|c c c|c c c|c c c|}
		\hline
		\textbf{Category} & \multicolumn{3}{c|}{\textbf{Company}} & \multicolumn{3}{c|}{\textbf{Job Title}} & \multicolumn{3}{c|}{\textbf{School}} & \multicolumn{3}{c|}{\textbf{Skill}} \\ \hline
		\textbf{Metric} & {\it P}  & {\it R} & $F1$ & {\it P}  & {\it R} & $F1$ & {\it P}  & {\it R} & $F1$ & {\it P}  & {\it R} & $F1$      \\ \hline
		$bow$               &    91.46 &   79.72  &  85.19  &  84.92  &  90.08  &  87.42 & 99.07  &  94.23  &  96.59 & 66.07  &  91.04  &  76.57  \\ \hline
		$wiki_w$              &   88.92  &  92.23 &   90.54 & 85.85  &  93.82  &  \textbf{89.66} & 98.92  &  96.42  &  97.66  & 87.36  &  88.15  &  87.75   \\ \hline
		$wiki_x$              &       95.41   & 96.55 &   \textbf{95.98}  & 86.27  &  88.30  &  87.28  &  98.93   & 98.11  &  \textbf{98.52} & 91.99  &  92.42  &  \textbf{92.21}         \\ \hline
		
	\end{tabular}
\end{table*}

\begin{table*}[]
	\centering
	\caption{Performance of Different \textit{ETR} Models on the Labeled Entity Data set using Ensemble of Features}
	\label{my-label}
	\begin{tabular}{|l|c c c|c c c|c c c|c c c|}
		\hline
		\textbf{Category} & \multicolumn{3}{c|}{\textbf{Company}} & \multicolumn{3}{c|}{\textbf{Job Title}} & \multicolumn{3}{c|}{\textbf{School}} & \multicolumn{3}{c|}{\textbf{Skill}} \\ \hline
		\textbf{Metric} & {\it P}  & {\it R} & $F1$ & {\it P}  & {\it R} & $F1$ & {\it P}  & {\it R} & $F1$ & {\it P}  & {\it R} & $F1$      \\ \hline
		$wiki_x$              &       95.41   & 96.55 &   95.98  & 86.27  &  88.30  &  87.28  &  98.93   & 98.11  &  98.52 & 91.99  &  92.42  &  92.21         \\ \hline
		%$job_w$              &   91.66      &     81.95     &     86.54    &    83.43 &   93.24  &  88.06 & 98.77  &  94.94  &  96.81 & 69.81 &   90.53  &  78.83         \\ \hline		
		$wiki_x,job_w$         &  95.64  &  96.73  &  96.18  &   88.32  &  91.99 &   90.12   &   99.16  &  98.20  &  98.68    &  92.45   & 93.17   & 92.81     \\ \hline
		$wiki_x,job_w,ont$     &   96.38 &   96.72  &  96.55  & 87.94  &  92.13 &   89.98  & 99.14 &   98.25  &  98.69 &  92.33  &  93.95  &  93.13     \\ \hline
		$wiki_x,job_w,lex$     &   95.67  &  96.68  &  96.17 & 88.34  &  92.82  &  90.53 & 99.16 &   98.21  &  98.68 & 92.49  &  93.14  &  92.81   \\ \hline
		$wiki_x,job_w,lex,ont$ &  96.41  &  96.72  &  \textbf{96.56}   &  88.35  &  92.91 &   \textbf{90.57} &  99.15  &  98.23  &  \textbf{98.69}  &  92.31 &   93.99  &  \textbf{93.14}  \\ \hline
		
	\end{tabular}
\end{table*}

\section{Experiments and Results}
In this section we present our empirical results. We start by describing the data set used in experiments and then detail different models developed for \textit{ETR} along with their results. 
\subsection{Data set}
We build our \textit{ETR} models using the largest labeled entity data set owned by CareerBuilder. The data set contains more than 177K labeled entities distributed over four categories as shown in Table \uppercase\expandafter{\romannumeral3\relax}. These entities were obtained from CareerBuilder's search logs, job postings, and resume postings, and were manually reviewed by annotators working at CareerBuilder. 

\subsection{Experimental Setup}
We conducted several experiments in order to evaluate the performance of the \textit{ETR} system with different models. We started by evaluating models built from a single feature source i.e., contextual vectors or word synonymy vectors. Then we evaluated a model built using an ensemble of both of these distributional vectors. Finally, we evaluated a model which combines both distributional vectors plus the entity's ontological and lexical features (i.e., $ont_{e_j}$ and $lex_{e_j}$ respectively). 

To assess the effectiveness of our approach, we built two baseline models. The first one is the bag-of-words ($bow$) model which depends solely on words that appear in search entities as features without any contextual enrichment. The second model ($wiki_w$) is a distributional semantic model built by training word2vec on Wikipedia. After word2vec produces word distributional vectors, word synonymy vectors of search entities are generated as described in Section \uppercase\expandafter{\romannumeral3\relax}.D. We then generate a tf-idf entity-word matrix from these vectors as described in Section \uppercase\expandafter{\romannumeral3\relax}.G.

We built the Wikipedia search index using the English \textit{Wikipedia} dump of March 2015\footnote{\url{https://dumps.wikimedia.org/enwiki/20150304/}}. The total uncompressed XML dump size was about 52GB representing about 7 million articles. We extracted the articles using a modified version of the Wikipedia Extractor\footnote{http://medialab.di.unipi.it/wiki/Wikipedia\textunderscore Extractor}. Our version\footnote{https://github.com/walid-shalaby/wikiextractor} extracts articles as plain text, discarding images and tables. We discarded the {\it References} and {\it External Links} sections (if any). We pruned all articles which are not under the main namespace, and excluded all \textit{disambiguation}, \textit{list of}, and \textit{redirect} pages as well. Eventually, our index contained about 4 million documents.

While searching the Wikipedia index, we search both content and title fields. For efficiency, we limit retrieved results to the top 3 hits which have a minimum length of 100 bytes.

To build the word embedding vectors, we trained word2vec on more than 60 million job postings from CareerBuilder. We used Apache Spark's scalable machine learning library (MLlib\footnote{https://spark.apache.org/mllib/}) which has an implementation of word2vec in Scala\footnote{http://www.scala-lang.org/}. We configured the parameters of the word2vec model as follows: minimum word count = 50, number of iterations (epoch)=1, vector size = 300, and number of partitions = 5000. The model took about 32 hours to fit on one of CareerBuilder's Hadoop clusters with 69 data nodes, each having a 2.6 GHz AMD Opteron Processor with 12 to 32 cores and 32GB to 128GB RAM.

Finally, we evaluate all the \textit{ETR} models using a Support Vector Machine (SVM) classifier with a linear kernel, leveraging the scikit-learn machine learning library \cite{scikit-learn}. Because entity instance frequencies over categories is a bit skewed and to avoid overfitting, we configured the classifier to use a different regularization value for each category relative to category frequencies. For each model we report Precision (\textit{P}), Recall (\textit{R}), and their harmonic mean (\textit{F1}) scores. All results are calculated using 10-fold cross-validation over the labeled entities data set. Folds were randomly generated using stratified sampling.

\subsection{Results}
Table \uppercase\expandafter{\romannumeral4\relax} shows the results obtained from the baseline models compared to the contextual vectors model using 10-fold cross-validation on the labeled entities data set. 

The first baseline model is the \textit{bow}. This model gives relatively lower \textit{F1} scores on all categories as shown in Table \uppercase\expandafter{\romannumeral4\relax}. Due to the absence of contextual information, this model fails to generalize well with unseen entities, as they contain terms that are not in the model's feature space. This is very clear with categories that have high naming variations (i.e., {\it Company} and {\it Skill}). {\it bow} performs relatively well on {\it Job Title} as it has limited naming variations. It also performs very well on {\it School} entities as they have common naming conventions (e.g., university, school, institute...etc). 

The second baseline model is $wiki_w$ which is built by training word2vec on Wikipedia. This model utilizes contextual features inferred from word distributional properties, hence it performs better than \textit{bow} on all categories. As shown in Table \uppercase\expandafter{\romannumeral4\relax}, the $wiki_w$ \textit{F1} score is higher than $bow$ by more than 5\% on \textit{Company}, 2\% on \textit{Job Title}, 1\% on {\it School}, and 11\% on \textit{Skill}. Those results indicate the viability of distributional semantic representations for {\it ETR} of short search entities.

The third model is $wiki_x$ which is built using contextual vectors generated by searching the Wikipedia index. It retrieves search entity contexts and category information from search hits and then utilizes them as learning features. As shown in Table \uppercase\expandafter{\romannumeral4\relax}, this novel approach outperforms both $bow$ and $wiki_w$ models substantially on {\it Company} and {\it Skill}. It also performs slightly better on {\it School}. These results indicate the effectiveness of the $wiki_x$ model in recognizing these categories accurately.

It is important to mention that, though both the $wiki_x$ and $wiki_w$ models use Wikipedia as an intermediate corpus to learn distributional representations of words, the $wiki_x$ representations are more successful for the {\it ETR} task. Compared with the $wiki_w$ model, the \textit{F1} scores of the $wiki_x$ model increased on the \textit{Company} class by 5\%, on the \textit{School} class by 1\%, and on the \textit{Skill} class by 5\%. 

The \textit{Job Title} category is the only example where the $wiki_w$ model performed better (by 2\%) than the $wiki_x$ model. A closer look at the scores reveals that, the $wiki_x$ model is more accurate than the $wiki_w$ model as it has a higher {\it P} score. The $wiki_w$ model, however, has better coverage as it has a higher \textit{R} score. Considering the small size of the \textit{Job Title} category (\textasciitilde3,600 entities), that difference in recall cannot be considered substantial.

The results in Table \uppercase\expandafter{\romannumeral4\relax} prove empirically that, for {\it ETR} of search entities, our novel approach for modeling real-world knowledge using contextual distributional representations outperforms word2vec, the state-of-the-art for distributional semantic representations, even though both use the same intermediate corpus (Wikipedia). Moreover, our method is much simpler and more efficient than word2vec as it doesn't require optimizing an objective function for learning word embedding vectors.

In order to increase overall system performance, we built four \textit{ETR} models that combine features from different sources as described in Section \uppercase\expandafter{\romannumeral3\relax}.G. We first built $job_w$ which models domain-specific knowledge of search entities. The $job_w$ model is built by training word2vec on the textual content of millions of job postings.

As shown in Table \uppercase\expandafter{\romannumeral5\relax}, we combined both contextual vector ($wiki_x$) and synonyms vector ($job_w$) representations and built an ensemble of the two models ($wiki_x$,$job_w$). The ensemble improved the results over $wiki_x$ across all categories. the largest improvement was on \textit{Job Title}, which saw a 3\% improvement in $F1$ score. More importantly, this ensemble outperforms the $wiki_w$ and $bow$ models on all categories.   
%$job_w$ performs better than the $wiki_w$ only on the \textit{School} category (97\% vs. 96\%). On the other hand, $wiki_w$ performs slightly better on \textit{Company} (89\% vs. 87\%) and \textit{Job Title} (89\% vs. 88\%) categories and much better on the \textit{Skill} category (84\% vs. 79\%).

To further increase system accuracy on {\it Company} class, we incorporated the DBpedia ontological type of search entity ($ont$) with contextual and synonymy vectors as described in Section \uppercase\expandafter{\romannumeral3\relax}.E. This ensemble ($wiki_x$,$job_w$,$ont$), as shown in Table \uppercase\expandafter{\romannumeral5\relax}, increased $F1$ score on {\it Company} by \textasciitilde0.4\%.

The third ensemble is ($wiki_x$,$job_w$,$lex$). It aims at increasing system accuracy on {\it Job Title} class by incorporating entity's linguistic features ($lex$) as described in Section \uppercase\expandafter{\romannumeral3\relax}.F. As shown in Table \uppercase\expandafter{\romannumeral5\relax}, the $F1$ score on {\it Job Title} increased by \textasciitilde0.6\% when incorporating this feature.

Finally, we combined all features generating an ensemble of contextual vectors, synonymy vectors, ontological features, and linguistics features ($wiki_x,job_w,lex,ont$). As shown in Table \uppercase\expandafter{\romannumeral5\relax}, this model produced the best $F1$ scores on all categories among all the aforementioned models.

%why entity recognition in search queries are challenging (no context, short text)
%why gazzetters will fail
%how it is different from NER
%why typical approaches like (CRF) will fail 

%Wikipedia, the largest human knowledge base, 

%Because search queries are typically short (1-3 words), we need to collect representative and discriminative features that represent contextual infromation about the search entity. 

%simple, efficient model
%new feature set from wikipedia

% we don't hard code the ontological or lexical feature
\section{Conclusion}
In this paper we presented an effective approach for {\it ETR} of search query entities in the job search and recruitment domain. We proposed a novel ensemble of features which enrich short query entities with real-world and domain-specific knowledge. The ensemble entity representation model contains features representing: 1) contextual information in Wikipedia, 2) embedding information in millions of job postings, 3) class type in DBpedia for {\it Company} entities, and 4) linguistic properties in WordNet for {\it Job Title} entities. 

Our approach is novel and distinct from other {\it ETR} approaches. To our knowledge, generating distributional semantic vectors of query entities using contextual information from Wikipedia as a search index was not reported before in the literature.

Evaluation results on a data set of more than 177K search entities were very promising. The results showed that our Wikipedia-based model outperforms the state-of-the-art word2vec model trained on Wikipedia on three out of four target entity categories. Moreover, our ensemble representation could achieve 97\% micro-averaged $F1$ score on the four entity types outperforming the word2vec baseline by 6\% on {\it Company}, 1\% on {\it Job Title}, 1\% on {\it School}, and 5\% on {\it Skill}.

In terms of performance, our system takes 30ms per entity type request, making it efficient and appropriate for serving online search queries.

Our system has been integrated within CareerBuilder's semantic search engine, which improved the quality of search results for tens of millions of job seekers every month.

\bibliographystyle{ieeetr}
\bibliography{main}

\begin{thebibliography}{10}

\bibitem{kazama2007exploiting}
J.~Kazama and K.~Torisawa, ``Exploiting wikipedia as external knowledge for
  named entity recognition,'' in {\em Proceedings of the 2007 Joint Conference
  on Empirical Methods in Natural Language Processing and Computational Natural
  Language Learning (EMNLP-CoNLL)}, pp.~698--707, 2007.

\bibitem{nadeau2007survey}
D.~Nadeau and S.~Sekine, ``A survey of named entity recognition and
  classification,'' {\em Lingvisticae Investigationes}, vol.~30, no.~1,
  pp.~3--26, 2007.

\bibitem{minkov2005extracting}
E.~Minkov, R.~C. Wang, and W.~W. Cohen, ``Extracting personal names from email:
  Applying named entity recognition to informal text,'' in {\em Proceedings of
  the conference on Human Language Technology and Empirical Methods in Natural
  Language Processing}, pp.~443--450, Association for Computational
  Linguistics, 2005.

\bibitem{shaalan2007person}
K.~Shaalan and H.~Raza, ``Person name entity recognition for arabic,'' in {\em
  Proceedings of the 2007 Workshop on Computational Approaches to Semitic
  Languages: Common Issues and Resources}, pp.~17--24, Association for
  Computational Linguistics, 2007.

\bibitem{zhou2002named}
G.~Zhou and J.~Su, ``Named entity recognition using an hmm-based chunk
  tagger,'' in {\em proceedings of the 40th Annual Meeting on Association for
  Computational Linguistics}, pp.~473--480, Association for Computational
  Linguistics, 2002.

\bibitem{yin2010building}
X.~Yin and S.~Shah, ``Building taxonomy of web search intents for name entity
  queries,'' in {\em Proceedings of the 19th international conference on World
  wide web}, pp.~1001--1010, ACM, 2010.

\bibitem{aljadda2014pgmhd}
K.~AlJadda, M.~Korayem, C.~Ortiz, T.~Grainger, J.~A. Miller, and W.~S. York,
  ``Pgmhd: A scalable probabilistic graphical model for massive hierarchical
  data problems,'' in {\em Big Data (Big Data), 2014 IEEE International
  Conference on}, pp.~55--60, IEEE, 2014.

\bibitem{aljadda2015improving}
K.~AlJadda, M.~Korayem, and T.~Grainger, ``Improving the quality of semantic
  relationships extracted from massive user behavioral data,'' in {\em Big Data
  (Big Data), 2015 IEEE International Conference on}, pp.~2951--2953, IEEE,
  2015.

\bibitem{korayem2015query}
M.~Korayem, C.~Ortiz, K.~AlJadda, and T.~Grainger, ``Query sense disambiguation
  leveraging large scale user behavioral data,'' in {\em Big Data (Big Data),
  2015 IEEE International Conference on}, pp.~1230--1237, IEEE, 2015.

\bibitem{tjong2003introduction}
E.~F. Tjong Kim~Sang and F.~De~Meulder, ``Introduction to the conll-2003 shared
  task: Language-independent named entity recognition,'' in {\em Proceedings of
  the seventh conference on Natural language learning at HLT-NAACL 2003-Volume
  4}, pp.~142--147, Association for Computational Linguistics, 2003.

\bibitem{ratinov2009design}
L.~Ratinov and D.~Roth, ``Design challenges and misconceptions in named entity
  recognition,'' in {\em Proceedings of the Thirteenth Conference on
  Computational Natural Language Learning}, pp.~147--155, Association for
  Computational Linguistics, 2009.

\bibitem{carmel2014erd}
D.~Carmel, M.-W. Chang, E.~Gabrilovich, B.-J.~P. Hsu, and K.~Wang, ``Erd'14:
  entity recognition and disambiguation challenge,'' in {\em ACM SIGIR Forum},
  vol.~48, pp.~63--77, ACM, 2014.

\bibitem{moro2015semeval}
A.~Moro and R.~Navigli, ``Semeval-2015 task 13: Multilingual all-words sense
  disambiguation and entity linking,'' {\em Proceedings of SemEval-2015}.

\bibitem{ren2015clustype}
X.~Ren, A.~El-Kishky, C.~Wang, F.~Tao, C.~R. Voss, and J.~Han, ``Clustype:
  Effective entity recognition and typing by relation phrase-based
  clustering,'' in {\em Proceedings of the 21th ACM SIGKDD International
  Conference on Knowledge Discovery and Data Mining}, pp.~995--1004, ACM, 2015.

\bibitem{milne2008learning}
D.~Milne and I.~H. Witten, ``Learning to link with wikipedia,'' in {\em
  Proceedings of the 17th ACM conference on Information and knowledge
  management}, pp.~509--518, ACM, 2008.

\bibitem{cucerzan2007large}
S.~Cucerzan, ``Large-scale named entity disambiguation based on wikipedia
  data.,'' in {\em EMNLP-CoNLL}, vol.~7, pp.~708--716, 2007.

\bibitem{mansouri2008named}
A.~Mansouri, L.~S. Affendey, and A.~Mamat, ``Named entity recognition
  approaches,'' {\em International Journal of Computer Science and Network
  Security}, vol.~8, no.~2, pp.~339--344, 2008.

\bibitem{guo2009named}
J.~Guo, G.~Xu, X.~Cheng, and H.~Li, ``Named entity recognition in query,'' in
  {\em Proceedings of the 32nd international ACM SIGIR conference on Research
  and development in information retrieval}, pp.~267--274, ACM, 2009.

\bibitem{kulkarni2009collective}
S.~Kulkarni, A.~Singh, G.~Ramakrishnan, and S.~Chakrabarti, ``Collective
  annotation of wikipedia entities in web text,'' in {\em Proceedings of the
  15th ACM SIGKDD international conference on Knowledge discovery and data
  mining}, pp.~457--466, ACM, 2009.

\bibitem{han2011collective}
X.~Han, L.~Sun, and J.~Zhao, ``Collective entity linking in web text: a
  graph-based method,'' in {\em Proceedings of the 34th international ACM SIGIR
  conference on Research and development in Information Retrieval},
  pp.~765--774, ACM, 2011.

\bibitem{han2012entity}
X.~Han and L.~Sun, ``An entity-topic model for entity linking,'' in {\em
  Proceedings of the 2012 Joint Conference on Empirical Methods in Natural
  Language Processing and Computational Natural Language Learning},
  pp.~105--115, Association for Computational Linguistics, 2012.

\bibitem{lin2012entity}
T.~Lin, O.~Etzioni, {\em et~al.}, ``Entity linking at web scale,'' in {\em
  Proceedings of the Joint Workshop on Automatic Knowledge Base Construction
  and Web-scale Knowledge Extraction}, pp.~84--88, Association for
  Computational Linguistics, 2012.

\bibitem{richman2008mining}
A.~E. Richman and P.~Schone, ``Mining wiki resources for multilingual named
  entity recognition.,'' in {\em ACL}, pp.~1--9, 2008.

\bibitem{hu2009understanding}
J.~Hu, G.~Wang, F.~Lochovsky, J.-t. Sun, and Z.~Chen, ``Understanding user's
  query intent with wikipedia,'' in {\em Proceedings of the 18th international
  conference on World wide web}, pp.~471--480, ACM, 2009.

\bibitem{nothman2008transforming}
J.~Nothman, J.~R. Curran, and T.~Murphy, ``Transforming wikipedia into named
  entity training data,'' in {\em Proceedings of the Australian Language
  Technology Workshop}, pp.~124--132, 2008.

\bibitem{mohamedidentifying}
M.~Mohamed and M.~Oussalah, ``Identifying and extracting named entities from
  wikipedia database using entity infoboxes,'' {\em International Journal of
  Advanced Computer Science and Applications (IJACSA)}, vol.~5, no.~7, 2014.

\bibitem{gattani2013entity}
A.~Gattani, D.~S. Lamba, N.~Garera, M.~Tiwari, X.~Chai, S.~Das, S.~Subramaniam,
  A.~Rajaraman, V.~Harinarayan, and A.~Doan, ``Entity extraction, linking,
  classification, and tagging for social media: a wikipedia-based approach,''
  {\em Proceedings of the VLDB Endowment}, vol.~6, no.~11, pp.~1126--1137,
  2013.

\bibitem{laclavik2015search}
M.~Laclav{\'\i}k, M.~Ciglan, S.~Steingold, M.~Seleng, A.~Dorman, and
  S.~Dlugolinsky, ``Search query categorization at scale,'' in {\em Proceedings
  of the 24th International Conference on World Wide Web Companion},
  pp.~1281--1286, International World Wide Web Conferences Steering Committee,
  2015.

\bibitem{industrialPaper}
K.~AlJadda, M.~Korayem, T.~Grainger, and C.~Russell, ``Crowdsourced query
  augmentation through semantic discovery of domain-specific jargon,'' in {\em
  Big Data (Big Data), 2014 IEEE International Conference on}, pp.~808--815,
  IEEE, 2014.

\bibitem{mikolov2013efficient}
T.~Mikolov, K.~Chen, G.~Corrado, and J.~Dean, ``Efficient estimation of word
  representations in vector space,'' {\em arXiv preprint arXiv:1301.3781},
  2013.

\bibitem{scikit-learn}
F.~Pedregosa, G.~Varoquaux, A.~Gramfort, V.~Michel, B.~Thirion, O.~Grisel,
  M.~Blondel, P.~Prettenhofer, R.~Weiss, V.~Dubourg, J.~Vanderplas, A.~Passos,
  D.~Cournapeau, M.~Brucher, M.~Perrot, and E.~Duchesnay, ``Scikit-learn:
  Machine learning in {P}ython,'' {\em Journal of Machine Learning Research},
  vol.~12, pp.~2825--2830, 2011.

\end{thebibliography}

\end{document}